\documentclass[conference]{IEEEtran}
\usepackage{cite}
\usepackage{amsmath,amssymb,amsfonts}
\usepackage{algorithmic}
\usepackage{graphicx}
\usepackage{textcomp}
\usepackage{xcolor}
\usepackage{hyperref}
\usepackage{bm}
\usepackage{booktabs}
\usepackage{multirow}
\usepackage[normalem]{ulem}
\useunder{\uline}{\ul}{}
\usepackage{colortbl}
\usepackage{threeparttable}
\def\BibTeX{{\rm B\kern-.05em{\sc i\kern-.025em b}\kern-.08em
    T\kern-.1667em\lower.7ex\hbox{E}\kern-.125emX}}
\begin{document}

\title{Title}
\title{VE: Modeling Multivariate Time Series Correlation with Variate Embedding}

\author{
\IEEEauthorblockN{
    Shangjiong Wang\textsuperscript{1}, 
    Zhihong Man\textsuperscript{1}, 
    Zhenwei Cao\textsuperscript{1}, 
    Jinchuan Zheng\textsuperscript{1} and 
    Zhikang Ge\textsuperscript{2}}
\IEEEauthorblockA{
\textit{\textsuperscript{1}School of Science, Computing and Engineering Technologies} \\
\textit{Swinburne University of Technology}, VIC 3122, Australia \\
\textit{\textsuperscript{2}ZJU-Hangzhou Global Scientific and Technological Innovation Center} \\
\textit{Zhejiang University}, Hangzhou, China \\
\textsuperscript{1}\{shangjiongwang,zman,zcao,jzheng\}@swin.edu.au  
\textsuperscript{2}zge@zju.edu.cn  
}
}

\maketitle
\begin{abstract}
    Multivariate time series forecasting relies on accurately capturing the correlations among variates. 
    Current channel-independent (CI) models and models with a CI final projection layer are unable to capture these dependencies. 
    In this paper, we present the variate embedding (VE) pipeline, which learns a unique and consistent embedding for each variate and combines it with Mixture of Experts (MoE) and Low-Rank Adaptation (LoRA) techniques to enhance forecasting performance while controlling parameter size. 
    The VE pipeline can be integrated into any model with a CI final projection layer to improve multivariate forecasting. 
    The learned VE effectively groups variates with similar temporal patterns and separates those with low correlations. 
    The effectiveness of the VE pipeline is demonstrated through experiments on four widely-used datasets.
    The code is available at: \href{https://github.com/swang-song/VE}{https://github.com/swang-song/VE}.



\end{abstract}


\section{Introduction}
Multivariate time series forecasting (MTSF) is of great importance, as many downstream applications, such as energy consumption management and weather forecasting, rely on the accurate extrapolation of current multivariate data into the future. 
Multivariate correlations play a crucial role in producing more accurate forecasts. For instance, clients with similar lifestyles may exhibit similar electricity consumption patterns, while those with active nightlife may have different consumption patterns compared to individuals who prefer staying at home. 
Therefore, capturing and utilizing highly correlated variates in forecasting while minimizing the impact of low-correlation variates is a critical task.

To handle multivariate time series, two methods are commonly used to embed the input series: channel projection embedding and temporal projection embedding. 
Channel projection embedding projects all the variates at each time point into a representation, but this approach often leads to inferior performance due to the low semantic meaning at each time point and the differing physical meanings of each variate \cite{liu2023itransformer}. 
In contrast, temporal projection embedding preserves the temporal order of a sequence and projects temporal points into a representation for each variate, achieving better empirical performance \cite{zeng2023are, nie2022time}.

After obtaining the temporal projection embedding, there are channel-independent (CI) models that treat all variates the same \cite{zeng2023are, nie2022time} and attention models that leverage the attention mechanism \cite{vaswani2017attention} to exploit multivariate dependencies \cite{liu2023itransformer, ilbert2024unlocking}. 
However, both methods have limitations.
CI models, due to weight sharing among all variates, can only learn general patterns and cannot model patterns unique to specific variates as the distinct pattern in one variate is a noise for other variates with low correlation.
Meanwhile, attention models fail to provide a unified and consistent representation for each variate, as the attention maps evolve across heads and layers and become more abstract in deeper layers. 
Additionally, even with the attention mechanism, the final projection layer remains channel-independent, which is unable to react to variate-specific patterns.
Therefore, we pose the question: 
\textbf{Can we train a unique and consistent representation for each variate and utilize this in CI models or layers to improve forecasting?}


In this paper, we propose a variate embedding (VE) pipeline that trains a unique and consistent embedding for each variate to facilitate multivariate forecasting.
Recognizing that many state-of-the-art (SOTA) models use a CI linear projection layer as the final layer, we leverage VE (after applying a softmax operation) as weights to a set of candidate linear weights. 
These weighted weights are then used in the final linear layer, enabling the model to learn the optimal projection weights for each variate.
This approach can be interpreted as a mixture-of-experts (MoE) architecture \cite{shazeer2017outrageously} applied to the last linear layer, with the variate embedding serving as the gating mechanism.
To maintain parameter efficiency, we further apply low-rank adaptation (LoRA) \cite{hu2021lora} to the candidate linear weights, which factorizes the linear weights into two sub-weights of lower rank. 
By combining MoE and LoRA, our VE pipeline effectively captures multivariate correlations and enhances forecasting performance, especially when the variates exhibit highly diverse patterns.


To sum up, the contributions of this work include:
\begin{itemize}
    \item We propose variate embedding (VE), a new paradigm for multivariate time series modeling, on top of the existing temporal projection embedding and channel projection embedding methods. 
    \item We propose the VE pipeline to use VE in combination with MoE and LoRA, specifically targeting the final linear projection layer, making our method applicable to all models where the last projection layer is CI.
    \item We conduct experiments of VE on four commonly used datasets, demonstrating its effectiveness in modeling multivariate correlations and improving forecasting while maintaining parameter efficiency.
\end{itemize}

\section{Related Works} 

\textbf{Time series analysis} is a long-standing problem, and the methods used to approach this problem have evolved from traditional statistical methods and basic deep neural network architectures to recent linear models and large foundation models.
Recently, inspired by large language models (LLMs) that handle a variety of natural language processing (NLP) tasks using a single foundation model, many works have aimed to develop a foundation model for time series \cite{gruver2023large, jin2023time, das2023decoder}. However, these models are often very large, requiring significant resources for both training and inference. In contrast to the trend of increasing model complexity, another line of work focuses on reducing model parameter size.
Following the findings in DLinear \cite{zeng2023are}, which showed that simple linear models can outperform sophisticatedly designed Transformer models in time series forecasting, many recent works have employed pure multilayer perceptron (MLP) models \cite{yi2023frequency, liu2023koopa, das2023long, chen2023tsmixer, xu2024fits, lin2024sparsetsf}. 
Notably, FITS \cite{xu2024fits} and SparseTSF \cite{lin2024sparsetsf} have reduced the model parameter size to as low as 10K and 1K, respectively, while maintaining SOTA performance in time series forecasting.

However, it should be noted that these linear models (DLinear, FITS, SparseTSF) are all channel-independent, meaning they are unable to model multivariate correlations and variate-specific patterns. 
Our proposed VE can enhance these linear models by providing them with variate-dependent projection weights.
Additionally, although prompt embedding in large foundation models \cite{jin2023time} also involves an embedding for different datasets, it is fundamentally different from the VE proposed in this paper. 
Firstly, prompt embedding are fixed and obtained using a frozen LLM, whereas we train a new embedding for each variate. 
Secondly, prompt embedding can only be applied along with LLMs for handling natural language, whereas our VE can be integrated into any model with a channel-independent final linear projection layer, including linear models, which prompt embedding cannot.


\textbf{Mixture-of-Expert (MOE)} layers consist of multiple sub-networks and each input is processed by an input-dependent combination of these experts \cite{shazeer2017outrageously}.
MoE has been utilized in Fedformer \cite{zhou2022fedformer} for adaptively selecting the pooling filters to extract trend from time series.
Moreover, MoLE \cite{ni2024mixture} applied MoE to the DLinear \cite{zeng2023are} and the experts are time-stamp dependent.

Although MoLE is perhaps the most similar work to our proposed VE pipeline, there are several fundamental differences:
\begin{itemize}
    \item Model Integration: MoLE applies MoE at the model level by replicating multiple instances of the original model, making it applicable only to linear models and prohibitively expensive to extend to larger models. In contrast, our VE pipeline can be applied to any model where the final linear projection layer is CI.
    \item Dependency Focus: MoLE lacks the ability to model multivariate correlations because its experts are time-stamp dependent, whereas our experts are variate-dependent, enabling our VE pipeline to capture variate-specific patterns and correlations.
    \item Parameter Efficiency: Our VE pipeline integrates LoRA to enhance parameter efficiency, which MoLE does not. As shown in our experiments, LoRA significantly reduces the number of parameters required to achieve the same forecasting accuracy.
\end{itemize}

Furthermore, our experiments will demonstrate that VE functions similarly to word embeddings in NLP, where time series with similar patterns are grouped together, while distinct patterns are separated.


\textbf{Low Rank Adaptation (LoRA)} was proposed to factorize weight matrices into two low-rank matrices, significantly reducing trainable parameters during the fine-tuning phase of LLMs \cite{hu2021lora}.
The idea of matrix factorization has been previously applied to MTSF for decomposing each time series into factors and basis. 
Specifically, TRMF \cite{yu2016temporal} applied temporal regularization to the bases using an autoregressive (AR) model, while Basisformer \cite{nibasisformer} used a smoothness loss to regularize the learned bases and contrastive learning to align factors for backcast and forecast.

However, these methods focus on learning factorized temporal bases, while our VE factorizes the weight parameters as proposed in LoRA. Additionally, our VE can learn variate-specific features that are beneficial for improving the accuracy of MTSF.



\begin{figure*}[!t]
    \centering
    \includegraphics[width=0.9\linewidth]{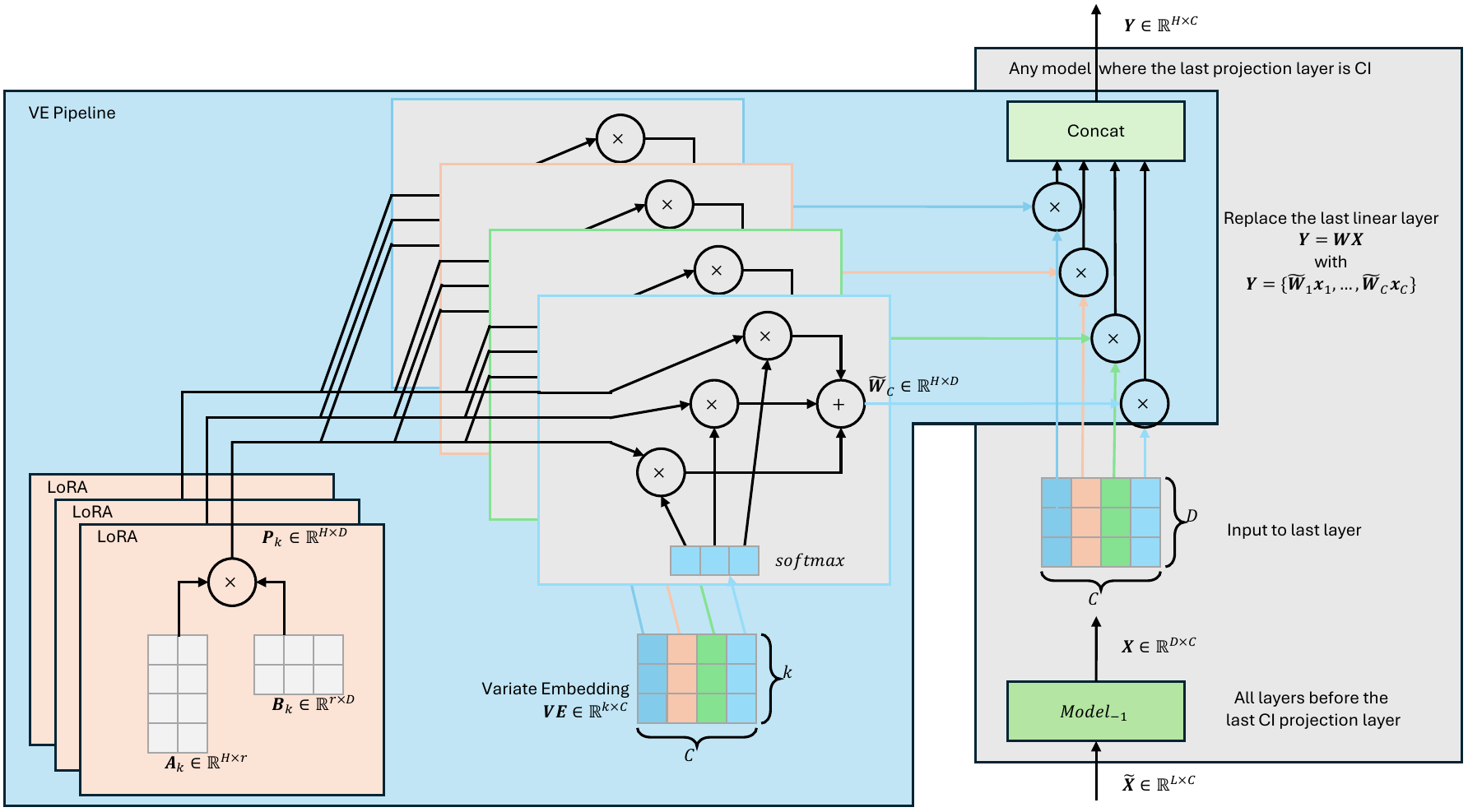}
    \caption{Overview of the proposed method. 
    Our VE pipeline can be integrated into any model where the final projection layer is channel-independent. 
    In our VE pipeline, we generate an embedding of dimension $k$ for each variate. Each embedding is then processed by a $softmax$ function, and the results are used as weights for a set of linear weights, which are decomposed into two low-rank matrices to reduce the parameter size. 
    The weighted linear weights are then used to do projection for each variate, replacing the original channel-independent weights.
    }
    \label{fig::VE_pipeline}
\end{figure*}

\section{VE Pipeline} 

The task of multivariate time series forecasting is to predict future time series 
$\tilde{\bm{Y}} \in \mathbb{R}^{H \times C}$ given the past time series 
$\tilde{\bm{X}} \in \mathbb{R}^{L \times C}$, where
$C$ is the number of variates to be predicted simultaneously, 
$H$ is the predict horizon and
$L$ is the look-back window length.
As stated in the introduction, our VE targets models where the final projection layer is channel-independent. 
Suppose the model dimension is 
$D$,
with the last layer input 
$\bm{X} = \{\bm{x}_{1}, \dots, \bm{x}_{C}\} \in \mathbb{R}^{D \times C}$ and 
the last layer weights
$\bm{W} \in \mathbb{R}^{H \times D}$, 
the prediction is formulated as
$\bm{Y} = \bm{W}\bm{X} = \{\bm{W}\bm{x}_{1}, \dots, \bm{W}\bm{x}_{C}\}$, where 
$\bm{Y} \in \mathbb{R}^{H \times C}$.

For linear models, the final layer input 
$\bm{X} = \tilde{\bm{X}}$, and 
for models with multiple layers, the final layer input is
$\bm{X} =  Model_{-1}(\tilde{\bm{X}})$, where 
$Model_{-1}$ represents the model excluding the last CI projection layer.

Due to weight sharing in the last projection layer, the weights
$\bm{W}$ can only learn general patterns common to all variates. 
It is unable to capture variate-specific patterns, as unique patterns in one variate are treated as noise for another variate with low correlation. Moreover, multivariate correlations are completely ignored due to weight sharing.


To enable variate-specific weights, each variate needs its own weights to encourage variate-specific responses. 
However, naively implementing individual weights not only dramatically increases model complexity when $C$ is large and leads to overfitting, but also lacks the exploration multivariate correlation, which is crucial for accurate MTSF.
These downsides have contributed to the prevalence of CI models like DLinear \cite{zeng2023are} and PatchTST \cite{nie2022time} in the literature. 


\subsection{VEMoE}
To address this issue, we first propose variate embedding $\bm{VE} \in \mathbb{R}^{k \times C}$, a look-up table to model multivariate correlations. 
Then, considering that for variates that share similar patterns, the weights specialized for that pattern should be reused across all relevant variates, while each variate should also attend to weights specialized for its specific pattern. 
To achieve this, we implement MoE to train $k$ linear experts $\bm{P} \in \mathbb{R}^{H \times D \times k}$, using the variate embedding after the softmax operation as the weights for the experts. This is formulated as:


\begin{equation}
    \tilde{\bm{W}} = \bm{P} \bm{W}_{ve}
\end{equation} 
where 
$\tilde{\bm{W}} \in \mathbb{R}^{H \times D \times C}$,
$\bm{W}_{ve} = softmax(\bm{VE})$ and 
$softmax$ is applied collumn-wise.
The final projection is given by:
\begin{equation}
    \bm{Y} = \{\tilde{\bm{W}}_{1}\bm{x}_{1}, \dots, \tilde{\bm{W}}_{C}\bm{x}_{C}\}
\end{equation} 
where 
$\tilde{\bm{W}}_{i}  \in \mathbb{R}^{H \times D}$ for 
$i = 1, \dots, C$.

In this way, the model complexity becomes manageable as it is controlled by the number of experts 
$k$ rather than the number of variates
$C$.
Also, variate-dependent linear weights is achieved as each variate now can attend to experts that specialize in its specific pattern.
Moreover, since the variate embeddings represent weights to the experts, variates with similar patterns can learn similarl embedding weights for those experts.
This allows our VE to function like word embeddings in NLP, grouping variates with high correlations together while diffusing those with low dependencies apart.


\subsection{LoRA}

The parameter size for the original linear layer is $(D+1) \times H$, including the bias term. 
After applying VEMoE, the linear layer parameter size increases to $C \times k + k \times (D+1) \times H$. This shows that the model complexity increases linearly with $k$. As a large $k$ greatly increases both the computational cost and the risk of overfitting, we use LoRA to further reduce the parameter size. Each projection weight $\bm{P}_i \in \mathbb{R}^{H \times D}$ is decomposed into two low-rank matrices $\bm{A}_i \in \mathbb{R}^{H \times r}$ and $\bm{B}_i \in \mathbb{R}^{r \times D}$ with rank $r$ for $i = 1, \dots, k$. This is formulated as:
\begin{equation}
    \bm{P} = \{\bm{P}_1, \dots, \bm{P}_k\} = \{\bm{A}_1 \bm{B}_1, \dots, \bm{A}_k \bm{B}_k\}
\end{equation} 

The parameter size now is reduced to 
$C \times k + k \times r \times (D+1+H)$.
To better manage the model parameter size, we set 
$r$ as:
\begin{equation}
    r = \left\lfloor \frac{
        p \times (D+1) \times H
    }{
        k \times (D+1+H) 
    }
    \right\rfloor
\end{equation} 
where 
$p$ is the parameter size expansion ratio, and 
$\left\lfloor \cdot  \right\rfloor$ denotes the floor operation.
When 
$p$ is set to 4, the parameter size roughly increases by a factor of four.


\begin{table*}[htb]
    \centering
    \begin{threeparttable}
    \caption{Comparison of baselines and proposed VE on ETTh1, ETTh2, ECL and Weather with prediction horizon 
    $H \in \{92, 192\}$.
    The \textbf{best results} are highlighted in \textbf{bold}. 
    }
    \label{tab:mse_all}
    \begin{tabular}{@{}cccccccccccc@{}}
    \toprule
    \multicolumn{2}{c}{Method}                                               & FITS  & \textbf{+VE}                        & Linear & \textbf{+VE}                        & DLinear        & \textbf{+VE}                        & iTransformer   & \textbf{+VE}                        & PatchTST       & \textbf{+VE}   \\ \midrule
    \multicolumn{1}{c|}{\multirow{2}{*}{ETTh1}}   & \multicolumn{1}{c|}{96}  & 0.372 & \multicolumn{1}{c|}{\textbf{0.370}} & 0.385  & \multicolumn{1}{c|}{\textbf{0.379}} & 0.375          & \multicolumn{1}{c|}{\textbf{0.371}} & \textbf{0.383} & \multicolumn{1}{c|}{0.385}          & 0.387          & \textbf{0.380} \\
    \multicolumn{1}{c|}{}                         & \multicolumn{1}{c|}{192} & 0.409 & \multicolumn{1}{c|}{\textbf{0.404}} & 0.418  & \multicolumn{1}{c|}{\textbf{0.414}} & \textbf{0.406} & \multicolumn{1}{c|}{\textbf{0.406}} & \textbf{0.437} & \multicolumn{1}{c|}{\textbf{0.437}} & 0.432          & \textbf{0.428} \\ \midrule
    \multicolumn{1}{c|}{\multirow{2}{*}{ETTh2}}   & \multicolumn{1}{c|}{96}  & 0.272 & \multicolumn{1}{c|}{\textbf{0.270}} & 0.284  & \multicolumn{1}{c|}{\textbf{0.279}} & \textbf{0.276} & \multicolumn{1}{c|}{\textbf{0.276}} & 0.304          & \multicolumn{1}{c|}{\textbf{0.300}} & \textbf{0.291} & 0.292          \\
    \multicolumn{1}{c|}{}                         & \multicolumn{1}{c|}{192} & 0.331 & \multicolumn{1}{c|}{\textbf{0.327}} & 0.345  & \multicolumn{1}{c|}{\textbf{0.333}} & 0.341          & \multicolumn{1}{c|}{\textbf{0.331}} & 0.380          & \multicolumn{1}{c|}{\textbf{0.377}} & 0.361          & \textbf{0.357} \\ \midrule
    \multicolumn{1}{c|}{\multirow{2}{*}{ECL}}     & \multicolumn{1}{c|}{96}  & 0.135 & \multicolumn{1}{c|}{\textbf{0.131}} & 0.147  & \multicolumn{1}{c|}{\textbf{0.144}} & 0.140          & \multicolumn{1}{c|}{\textbf{0.134}} & 0.148          & \multicolumn{1}{c|}{\textbf{0.140}} & -              & -              \\
    \multicolumn{1}{c|}{}                         & \multicolumn{1}{c|}{192} & 0.150 & \multicolumn{1}{c|}{\textbf{0.147}} & 0.161  & \multicolumn{1}{c|}{\textbf{0.158}} & 0.154          & \multicolumn{1}{c|}{\textbf{0.149}} & 0.162          & \multicolumn{1}{c|}{\textbf{0.157}} & -              & -              \\ \midrule
    \multicolumn{1}{c|}{\multirow{2}{*}{Weather}} & \multicolumn{1}{c|}{96}  & 0.170 & \multicolumn{1}{c|}{\textbf{0.142}} & 0.175  & \multicolumn{1}{c|}{\textbf{0.147}} & 0.174          & \multicolumn{1}{c|}{\textbf{0.152}} & 0.177          & \multicolumn{1}{c|}{\textbf{0.169}} & 0.154          & \textbf{0.147} \\
    \multicolumn{1}{c|}{}                         & \multicolumn{1}{c|}{192} & 0.212 & \multicolumn{1}{c|}{\textbf{0.185}} & 0.218  & \multicolumn{1}{c|}{\textbf{0.191}} & 0.217          & \multicolumn{1}{c|}{\textbf{0.196}} & 0.228          & \multicolumn{1}{c|}{\textbf{0.219}} & 0.200          & \textbf{0.194} \\ \bottomrule
    \end{tabular}
    \begin{tablenotes}
      \item '-' denotes it reported an out-out-memory error.
    \end{tablenotes}
    \end{threeparttable}
\end{table*}

\section{Experiments} 

\subsection{Experimental Setup}

\textbf{Datasets}.
We evaluate our VE pipeline on four popular and publicly available real-world datasets:
ETTh1, ETTh2, ECL and Weather \cite{wu2021autoformer}. 


\textbf{Baseline models}.
The proposed VE pipeline can be integrated into any model where the final operation is a channel-independent linear projection. 
To demonstrate the effectiveness of our method across a variety of models, we incorporate it into three linear models and two Transformer models:
\begin{itemize}
    \item Linear \cite{zeng2023are}: This is used to demonstrate the basic setting where there is only one linear layer.
    
    \item DLinear \cite{zeng2023are}: This is used to demonstrate the case where there are multiple inputs with different projection weights in the last layer and the prediction is the sum of these projections.
    We treat these multiple weights as a group and train a single VE for all the weights within the same group, as as using separate VEs for different weights in a group did not yield statistically significant improvements and only increased the parameter size in our experiments.

    \item FITS \cite{xu2024fits}: This is used to show the compatibility with complex-value projection layers, beyond the standard real-value linear projection.
    
    \item PatchTST \cite{nie2022time}: This is used to demonstrate that our method is applicable to transformer models where attention is used to explore temporal dependencies of within a series and the weights are shared across all variates.
    
    \item iTransformer \cite{liu2023itransformer}: This is used to demonstrate the applicability to transformer models where attention is used to explore multivariate correlations, while maintaining a channel-independent final projection layer.
\end{itemize}

\textbf{Evaluation metirc}.
Consistent with previous works \cite{zeng2023are, liu2023itransformer, xu2024fits}, we use Mean Squared Error (MSE) as the primary metric to compare forecasting performance across models.

\subsection{Main Results}

\textbf{Multivariate forecasting for each dataset.}
The forecasting errors for ETTh1, ETTh2, ECL and Weather are presented in Table \ref{tab:mse_all}.
Our proposed VE frequently outperforms the baselines across all datasets. 
On the ECL and Weather datasets, VE consistently improves performance over all baselines for every prediction horizon. 
Notably, on the Weather dataset, VE improves over Linear and the recently proposed FITS by 16\% and 12\% at prediction horizons of 96 and 192, respectively. 
On ETTh1 and ETTh2, VE shows improvements in most cases, particularly for linear models, though the results are less consistent for transformer-based models.


\textbf{Multivariate forecasting for mixed dataset.}
The inconsistent results on ETTh1 and ETTh2 can be attributed to the limited diversity in patterns, as these datasets contain only 7 variates. 
To verify this, we train on a combined dataset comprising all four datasets to ensure a much richer diversity of patterns. 
We truncate the training, validation, and testing datasets of ECL and Weather to match the length of the corresponding ETT datasets.

Additionally, since the baselines have different default training hyperparameters, such as look-back length and training epochs, which significantly affect model performance, we standardize these by setting the look-back length to 360 and training epochs to 10, ensuring a fair comparison across the baselines on the mixed dataset.

The results are presented in Table \ref{tab:mix_mse}. 
Our VE achieves consistent improvements over all baselines in both horizon settings. Specifically, due to the exploration of multivariate correlations using the attention mechanism, none of the linear models surpass the transformer model iTransformer in all settings. 
However, after integration with VE, all linear models outperform iTransformer in every setting. 
Additionally, the performance of transformer model is further enhanced when applying VE.
This verifies the effectiveness of our proposed VE in modeling multivariate correlations, especially in datasets with diverse patterns.




\begin{table}[!ht]
    \centering
    \caption{
    Comparison of baselines and proposed VE on mixed dataset with prediction horizon 
    $H \in \{92, 192\}$.
    The \textbf{best results} are highlighted in \textbf{bold}. 
    }
    \label{tab:mix_mse}
    \begin{tabular}{@{}lcccc@{}}
    \toprule
    Prediction length                 & \multicolumn{2}{c}{96}                       & \multicolumn{2}{c}{192} \\ \cmidrule(lr){2-3} \cmidrule(lr){4-5}
    \multicolumn{1}{l|}{Last layer}   & original & \multicolumn{1}{c|}{\textbf{VE}}    & original & \textbf{VE}    \\ \midrule
    \multicolumn{1}{l|}{Linear}       & 0.160  & \multicolumn{1}{c|}{\textbf{0.152}} & 0.183  & \textbf{0.178} \\
    \multicolumn{1}{l|}{Dlinear}      & 0.158  & \multicolumn{1}{c|}{\textbf{0.151}} & 0.181  & \textbf{0.176} \\
    \multicolumn{1}{l|}{FITS}         & 0.159  & \multicolumn{1}{c|}{\textbf{0.148}} & 0.182  & \textbf{0.173} \\
    \multicolumn{1}{l|}{iTransformer} & 0.153  & \multicolumn{1}{c|}{\textbf{0.150}} & 0.181  & \textbf{0.180} \\ \bottomrule
    \end{tabular}
\end{table}

\begin{figure*}[!t]
    \centering
    \includegraphics[width=0.9\linewidth]{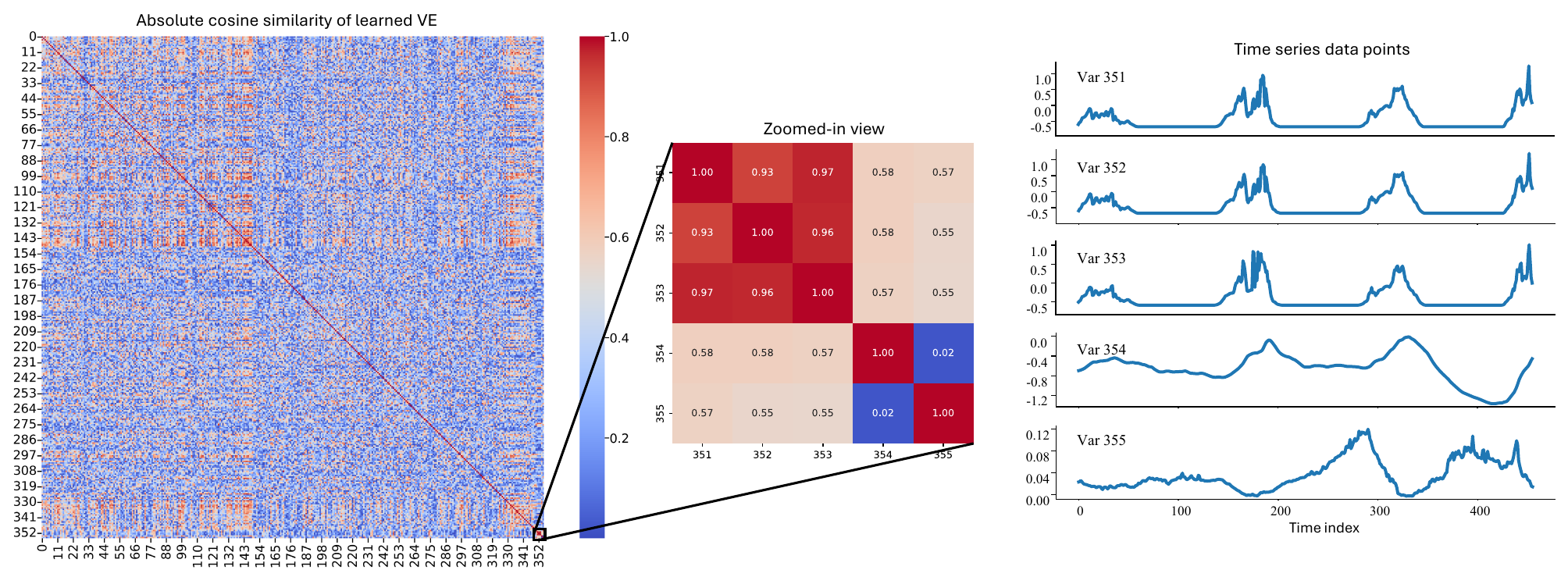}
    \caption{Visualization of VE similarity for FITS($H=96$) on mixed dataset.
    \textbf{Left:} absolute cosine similarity of the learned VE;
    \textbf{middle:} a zoomed-in view of the similaryty matrix for variates from 351 to 355;
    \textbf{right:} the data points of the five variates in the first sample of the first batch in the testing dataset.}
    \label{fig::VE_cos_similarity}
\end{figure*}

\begin{figure*}[!t]
    \centering
    \includegraphics[width=0.9\linewidth]{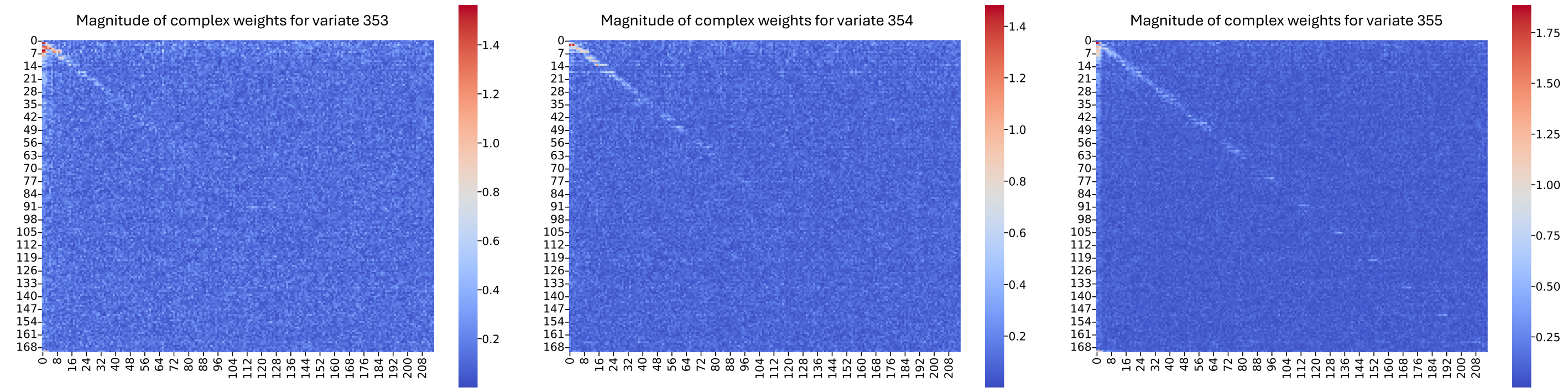}
    \caption{Visualization of the magnitude of the weighted complex-value wegiths of FITS($H=96$) for variates from 353 to 355.}
    \label{fig::VE_adaptiveProjection}
\end{figure*}

\subsection{Analysis of learned variate embedding}
To visualize the learned variate embedding (VE) and verify its ability to group variates with high correlations while diffusing others, we computed the cosine similarity of the VE among all 356 variates in the mixed dataset, taking the absolute value of these similarities. 
The complete absolute cosine similarity matrix is shown on the left of Fig. \ref{fig::VE_cos_similarity}. 
Notably, two distinct patterns are visible within the ECL dataset (variates 14 to 334), with a separation occurring around variate 149.

A zoomed-in view of the variates from 351 to 355 is presented in the middle panel of Fig. \ref{fig::VE_cos_similarity}, and their corresponding time points are plotted on the right. 
Variates 351, 352, and 353 exhibit similar temporal patterns, with cosine similarities exceeding 0.9, while the similarities for variates 354 and 355 approach zero (0.02), indicating distinct differences in their patterns.

Additionally, we visualized the magnitude of the weighted complex-value weights for the FITS model in Fig. \ref{fig::VE_adaptiveProjection}, focusing on variates 353 to 355. 
The weighted weights for variates 351 and 352 are omitted as they are nearly identical to those of variate 353. 
From variate 353 to 355, high-frequency information becomes increasingly important, with high-order harmonic frequencies of the dominant frequency 24 (corresponding to a base frequency of 15 in the figure) being particularly significant for variate 355.


These results confirm that VE effectively captures multivariate correlations, grouping variates with similar patterns together while differentiating less correlated variates.

\begin{table}[!t]
    \centering
    \caption{Ablation results of the proposed VE on ETTh2 ($H=192$).
    }
    \label{tab:Abalation}
    \begin{tabular}{@{}llcc@{}}
    \toprule
    \multicolumn{2}{l}{Method}                          & MSE$\downarrow$            & Parameters$\downarrow$      \\ \midrule
    \multicolumn{2}{l|}{FITS}                           & 0.331          & 48.86K          \\ \midrule
    \multicolumn{2}{l|}{FITS+VEMoE($k$=2)}                & 0.333          & 97.73K          \\
    \multicolumn{2}{l|}{FITS+VEMoE($k$=4)}                & 0.330          & 195.45K         \\
    \multicolumn{2}{l|}{FITS+VEMoE($k$=8)}                & 0.329          & 390.90K         \\
    \multicolumn{2}{l|}{FITS+VEMoE($k$=16)}               & {\ul 0.328}    & 781.81K         \\
    \multicolumn{2}{l|}{FITS+VEMoE($k$=32)}               & 0.329          & 1.56M           \\
    \multicolumn{2}{l|}{FITS+VEMoE($k$=64)}               & {\ul 0.328}    & 3.13M           \\
    \multicolumn{2}{l|}{FITS+VEMoE($k$=128)}              & 0.330          & 6.25M           \\ \midrule
    \multicolumn{2}{l|}{FITS+VEMoE+LoRA($k$=8,   $p$=0.25)} & 0.332          & \textbf{10.74K} \\
    \multicolumn{2}{l|}{FITS+VEMoE+LoRA($k$=32,   $p$=1)}   & {\ul 0.328}    & {\ul 42.94K}    \\
    \multicolumn{2}{l|}{FITS+VEMoE+LoRA($k$=128,   $p$=4)}  & \textbf{0.327} & 171.78K         \\ \bottomrule
    \end{tabular}
\end{table}

\subsection{Abalation study}
We conducted an ablation study on the two components, VEMoE and LoRA, of our VE pipeline using the FITS model on the ETTh2 dataset with a prediction horizon of 192, The results are presented in Table \ref{tab:Abalation}, with the \textbf{best results} in \textbf{bold} and the \underline{second best} results in \underline{underline}.

The default FITS model has 48.86K parameters and an MSE of 0.331. 
Our VEMoE alone improves forecasting but increases parameter size significantly. 
Adding LoRA reduces both the MSE and parameter size. 
Specifically, without LoRA, the improvement plateaus at an MSE of 0.328 with 16 times ($k=16$) the original FITS parameters. 
With LoRA, the MSE remains at 0.328, but with only 43K parameters ($p=1$). 
Remarkably, using just 10K parameters ($p=0.25$), the MSE of 0.332 is still competitive.


This study validates the effectiveness of VEMoE and LoRA in our VE pipeline, demonstrating that their combination provides optimal performance while maintaining parameter efficiency.

\section{Conclusion} 
This paper presents the VE pipeline, designed to capture multivariate correlations through the proposed variate embedding. 
By combining VE with a MoE architecture, the model can learn optimal weights for each variate, while low-rank adaptation (LoRA) is employed to improve parameter efficiency. 
Our VE pipeline is applicable to any model with a channel-independent final projection layer, addressing the limitation of such layers in learning unique patterns specific to each variate.
The effectiveness and compatibility of the proposed VE pipeline are validated through experiments on four widely-used datasets. 
We believe that our VE approach has the potential to be beneficial in other time series analysis tasks, such as time series imputation and anomaly detection, and building parameter-efficient time series foundation models to process various time series tasks and diverse data domains.



\bibliographystyle{IEEEtran}
\bibliography{icamechs2024_main}

\end{document}